\documentclass{article} 
\usepackage{iclr2016_conference,times}
\usepackage{hyperref}
\usepackage{url}
\usepackage{amssymb}
\usepackage{amsmath}
\usepackage{graphicx}
\usepackage{pbox}
\usepackage{tabu}

\title{An Information Retrieval Approach to Finding Dependent Subspaces of Multiple Views}

\author{Ziyuan Lin$^{1,*}$ \& Jaakko Peltonen$^{1,2,}$\thanks{Z. Lin and J. Peltonen contributed equally to the work.}\\
$^1$Helsinki Institute for Information Technology HIIT, Department of Computer Science,
Aalto University\\
$^2$School of Information Sciences, University of Tampere \\
\texttt{\{jaakko.peltonen,ziyuan.lin\}@aalto.fi} \\
}

\begin{document}

\maketitle

\begin{abstract}
Finding relationships between multiple views of data is essential 
both for exploratory analysis and as pre-processing for predictive tasks.
A prominent approach is to apply variants of Canonical Correlation Analysis (CCA), 
a classical method seeking correlated components between views.
The basic CCA is restricted to maximizing a simple dependency criterion, correlation, 
measured directly between data coordinates. 
We introduce a new method that finds dependent subspaces of views 
directly optimized for the
data analysis task of \textit{neighbor retrieval between multiple views}.
We optimize mappings for each view such as linear transformations to maximize
cross-view similarity between neighborhoods of data samples.
The criterion arises directly from the well-defined retrieval task,
detects nonlinear and local similarities, is able to measure dependency of data 
relationships rather than only individual data coordinates, and is related to
well understood measures of information retrieval quality.
In experiments we show the proposed method outperforms 
alternatives in preserving cross-view neighborhood similarities, 
and yields insights into local dependencies between multiple views.
\end{abstract}

\section{Introduction}

Finding dependent subspaces across views (subspaces where some property of the data is statistically strongly related or similar across views) can be useful as preprocessing for
predictive tasks if the non-dependent parts of each subspace may arise from
noise and distortions.
In some data analysis tasks, finding the dependent subspaces may itself be the
main goal; for example in bioinformatics domains dependency seeking projections
have been used to identify relationships between different views of cell
activity \citep{tripathi08, Klami13jmlr};
while in signal processing, a similar task could be identifying optimal filters for dependent signals of different nature \citep{xrmb}.

In the context of the more general multi-view learning \citep{chang13}, 
which learns models by leveraging multiple potentially dependent data views,
Canonical Correlation Analysis (CCA) \citep{HotellingCCA} is the standard unsupervised tool.
CCA iteratively finds component pairs that maximizes the correlations between the data points in the projected subspaces.
Correlation is a simple and restricted criterion to measure linear and global dependency.
To measure dependency in a more flexible way, particularly to handle nonlinear local dependency,
linear and nonlinear variants of CCA have been proposed.
For linear variants for this purpose,
Local CCA (LCCA) \citep{WeiX12LCCA} seeks linear projections for local patches in both views that maximizes correlation locally, 
and aligns the local linear projections into a global nonlinear projection. 
Its variant, Linear Local CCA (LLCCA) finds a linear approximation for the global nonlinear projection.
Locality Preserving CCA (LPCCA) \citep{SunC07LPCCA} maximizes a reweighted correlation between the differences 
of the data coordinates in both views. 
As a more general framework, 
Canonical Divergence Analysis \citep{hoangvuCDA} minimizes a general divergence measure between the probability distributions
of the data coordinates in the linearly projected subspace. 

The methods mentioned above work on data coordinates in the original spaces.
There are also nonlinear CCA variants (e.g., \cite{lai00, Bach2003, VerbeekRV03, 
galen2013-deep-cca, wang2015-deep-multi-view, hodosh2013})
for detecting nonlinear dependency between multiple views.
%
%
%
Although some of the above-mentioned variants are also locality-aware, they 
introduce the locality from the original space 
before maximizing the correlation or other similarity measures in the low-dimensional subspaces.
Since the locality in the original space may not necessarily reflect the locality in the subspaces,
such criteria may not suitable for finding dependencies that are local in the subspaces.

The methods discussed above, and several in \cite{chang13}, all either maximize correlation of data coordinates 
across views or alternative dependency measures between coordinates. We point out that
in many data domains the coordinates themselves may not be of main interest but rather the
data relationships that they reveal; it is then of great interest to develop dependency
seeking methods that directly focus on the data relationships.
In this paper we propose a method that does this: our method directly maximizes the \textit{between-view similarity of
neighborhoods of data samples}. It is a natural measure for similarity of data relationships among the views,
detects nonlinear and local dependencies,
and can be shown to be related to an information retrieval task of the analyst,
retrieving neighbors across views.
%
%
%
Our method is in principle general and suitable for finding both linear and
nonlinear data transformations for each view. Regardless of which kind of transformation is to be optimized,
the dependency criterion detects nonlinear dependencies across views; the type of transformation to be
optimized is simply
a choice of the analyst. In this first paper we focus on
and demonstrate the case of linear transformations which have the advantage of
simplicity and easy interpretability with respect to original data features;
however, we stress that this is not a limitation of the method.


\section{The Method: Dependent Neighborhoods of Multiple Views}

Our methodology focuses on analysis and preservation of neighborhood
relationships between views. We thus start by defining the neighborhood
relationships and then discuss how to measure their similarity across views.
We define the methodology for the general case of $N_{\textrm{Views}}>1$ views
but in experiments focus on the most common case of $N_{\textrm{Views}}=2$ 
views.

\subsection{Probabilistic Neighborhood Between Data Items}
Instead of resorting to a naive hard neighborhood criterion where two points
either are or are not neighbors, we will define a more realistic probabilistic
neighborhood relationship between data items. Any view that defines a feature
representation for data items can naturally be used to derive probabilistic
neighborhood relationships as follows.

Assume data items $x_i=(x_{i,1},\ldots,x_{i,N_{\textrm{Views}}})$ 
have paired features $x_{i,V}$ in each view $V$. 
We consider transformations of each view by a mapping $f_V$ which is typically
a dimensionality reducing transformation to a subspace of interest; in this paper, 
for simplicity
and interpretability we use linear mappings $f_V(x_{i,V})=W_V^\top x_{i,V}$ 
but the method is general and is not restricted to linear mappings.
The local neighborhood of a data item $i$ in any transformation of view $V$ can be represented by the
conditional probability distribution $p_{i,V}=\{p_V(j|i;f_V)\}$
where the $p_V(j|i;f_V)$ tell the probability that
another data item $j\neq i$ is picked as a representative neighbor of $i$; that
is, the probability that an analyst who has inspected item $i$ will next choose
$j$ for inspection. The probability $p_V(j|i;f_V)$ can be defined in several ways,
here we choose to define it through a simple exponential falloff with respect to
squared distance of $i$ and $j$, as
\begin{equation}
p_V(j|i;f_V) 
= \frac{\exp(-d_V^2(i,j;f_V)/\sigma_{i,V}^2)}
{\sum_{k\neq i} \exp(-d_V^2(i,k;f_V)/\sigma_{i,V}^2)} 
\label{eq:neighborhood_probability_general}
\end{equation}
where $d_V(i,j;f_V)$ is a distance function between the features of $i$ and $j$
in view $V$, and $\sigma_{i,V}$ controls the falloff rate around $i$ in the
view; in experiments we set $\sigma_{i,V}$ simply to a fraction of the maximum 
pairwise distance so that $\sigma_{i,V}=0.05\cdot\max_{j,k}||x_{j,V}-x_{k,V}||$ but
advanced local choices to e.g. achieve a desired entropy are possible, see for 
example \cite{Venna10jmlr}.
In the
case of linear mappings the probabilities become
\begin{equation}
p_V(j|i;f_V) 
= \frac{\exp(-(x_{i,V}-x_{j,V})^\top W_V W_V^\top (x_{i,V}-x_{j,V})/\sigma_{i,V}^2)}
{\sum_{k\neq i} \exp(-(x_{i,V}-x_{k,V})^\top W_V W_V^\top (x_{i,V}-x_{k,V})/\sigma_{i,V}^2)} \;.
\label{eq:neighborhood_probability_linear}
\end{equation}
where the transformation matrix $W_V$ defines the subspace of interest for the
view and also defines the distance metric within the subspace. Our method will
learn the mapping parameters (for linear mappings the matrix $W_V$) for each
view.

\subsection{Comparison of Neighborhoods Across Views}
\label{sec:comparing_neighborhoods}

When neighborhoods are represented as probabilistic distributions, they can be
compared using several difference measures that have been proposed between 
probability distributions. We discuss below two measures that will both be used
in out method for different purposes, and their information retrieval interpretations.

\textbf{Kullback-Leibler divergence.}
For two distributions $p=\{p(j)\}$ and $q=\{q(j)\}$, the Kullback-Leibler (KL) 
divergence is a well-known asymmetric measure of difference defined as
\begin{equation}
D_{KL}(p,q) = \sum_j p(j) \log \frac{p(j)}{q(j)} \;.
\end{equation}
The KL divergence is an information-theoretic criterion that is nonnegative and
zero if and only if $p=q$. Traditionally it is interpreted to measure the amount
of extra coding length needed when coding examples with codes generated for
distribution $q$ when the samples actually come from distribution $p$.
In our setting we will treat views symmetrically and compute the symmetrized
divergence $(D_{KL}(p,q)+D_{KL}(q,p))/2$. 

Importantly, the KL divergence $D_{KL}(p,q)$ can be shown to be
related to an \emph{information retrieval criterion}: cost of \emph{misses in
information retrieval of neighbors}, when neighbors following distribution $p$
are retrieved from a retrieval distribution $q$. The mathematical relationship
to information retrieval was shown by \cite{Venna10jmlr} where it was used for
comparing a reduced-dimensional neighborhood to an original one; here we use it
in a novel fashion to compare neighborhoods across (transformed) views of data.
The symmetrized divergence is thus the total cost of misses and false neighbors
from that neighbors following the neighbor distribution in one transformed view
are retrieved from the other transformed view with its neighbor distribution: 
the term $D_{KL}(p,q)$ is both the cost of false neighbors from that 
neighbors following $p$ are retrieved from $q$, and the cost of misses from 
that neighbors following $q$ are retrieved from $p$. 
Similarly, the term $D_{KL}(q,p)$ is both the cost of misses from that 
neighbors following $p$ are retrieved from $q$, and the cost of false neighbors from 
that neighbors following $q$ are retrieved from $p$.

While the KL divergence is a well-known difference measure its downside is that
its value can depend highly on differences between individual probabilities
$p(j)$ and $q(j)$, as a single missed neighbor can yield a very high value of the 
divergence: for any index $j$ if $p(j)>\epsilon$ for some $\epsilon>0$, 
$D_{KL}(p,q)\rightarrow \infty$ as $q(j)\rightarrow 0$. 
In real-life multi-view data differences between views may be unavoidable and
hence we prefer a less strict measure focusing more on overall similarity of the
neighborhoods than on severity of individual misses. We discuss such a measure
below.

\textbf{Angle cosine.}
An even simpler measure if similarity between two discrete probability 
distributions is the angle cosine between the distributions represented as 
vectors, that is,
\begin{equation}
\textrm{Cos}(p,q) = \frac{\sum_j p(j) q(j)}{\sqrt{(\sum_j (p(j))^2) (\sum_j (q(j))^2)}}
\end{equation}
The above angle cosine can also be interpreted as the Pearson correlation
coefficient between (unnormalized) vector elements of $p$ and $q$; it can thus
be seen as a neighborhood correlation, that is, a neighborhood based analogue of
the coordinate correlation cost function of CCA.\footnote{To make the connection
exact, typically correlation is computed after substracting the mean from
coordinates; for neighbor distributions in a data set of $n$ data items, the mean 
of neighborhood probabilities is the
data-independent value $1/(n-1)^2$ which can be substracted from each sum term
if an exact analogue to correlation is desired.} The angle cosine
measure is bounded from above and below: it attains the highest value $1$ if 
and only if $p=q$ and the lowest value $0$ if the supports of $p$ and $q$ are 
nonoverlapping.

\textbf{Similarity of neighborhoods by itself is not enough.}
The KL divergence and angle cosine (neighborhood correlation)
measures discussed above only compare
similarity of neighborhoods but not the potential usefulness of the found
subspaces where neighborhoods are similar. In high-dimensional data it is often
possible to find subspaces where neighborhoods are trivially similar. For
example, in data with sparse features it is often possible to find two
dimensions where all data is reduced to a single value; in such dimensions
neighborhood distributions would become uniform across all data since and hence
any two such dimensions would appear similar. To avoid discovering trivial
similarities we wish to complement the measures of similarity between
neighborhoods with terms that favor nontrivial (sparse) neighborhoods.

A simple way to prefer sparse neighborhoods is to omit the normalization from
the angle cosine (neighborhood correlation), yielding
\begin{equation}
\textrm{Sim}(p,q) = \sum_j p(j) q(j)
\end{equation}
which is simply the inner product between the vectors of neighborhood
probabilities. Unlike $\mathrm{Cos}(p,q)$, the measure $\mathrm{Sim}(p,q)$ favors sparse 
neighborhoods: it attains the highest value $1$ if and only if $p=q$ and 
$p(j)=q(j)=1$ for only a single element $j$, and the lowest value $0$ if the 
supports of $p$ and $q$ are nonoverlapping.

The information retrieval interpretation of $\mathrm{Sim}(p,q)$ is as follows: it is
a proportional count of true neighbors from $p$ retrieved from $q$ or vice versa. 
If $p$ has $K$ neighbors with near-uniform high probabilities
$p(j)\approx 1/K$ and other neighbors have near-zero probabilities, and $q$ has $L$ neighbors 
with high probability $q(j)\approx 1/L$, then $\mathrm{Sim}(p,q)\approx M/KL$ 
where $M$ is the number of neighbors for which both $p$ and $q$ 
have high probability (retrieved true neighbors). Hence $\mathrm{Sim}(p,q)$
rewards matching neighborhoods and favors
sparse neighborhoods where $K$ and $L$ are small.

\subsection{Final Cost Function and Technique for Nonlinear Optimization}

We wish to evaluate similarity of neighborhoods between subspaces of each view,
and optimize the subspaces to maximize the similarity, while at the same time
favoring subspaces containing sparse (informative) neighborhoods for data items.
We then evaluate the similarities as $\textrm{Sim}(p_{i,V},p_{i,U})$ where 
$p_{i,V}=\{p_V(j|i;f_V)\}$ is the neighborhood distribution around a data item $i$
in the dependent subspace of view $V$ and $f_V$ is the mapping (parameters) of the subspace, 
and $p_{i,U}=\{p_U(j|i;f_U)\}$ is the corresponding neighborhood distribution 
in the dependent subspace of view $U$ having the mapping $f_U$.

As the objective function for finding dependent projections, we sum the above 
over each pair of views $(U,V)$ and over the neighborhoods of each data 
item $i$, yielding
\begin{equation}
C(f_1,\ldots,f_{N_{\textrm{views}}}) 
= \sum_{V=1}^{N_{\textrm{views}}} \sum_{U=1,U\neq V}^{N_{\textrm{views}}} \sum_{i=1}^{N_{\textrm{data}}}
\sum_{j=1,j\neq i}^{N_{\textrm{data}}} p_V(j|i;f_V) p_U(j|i;f_U)
\label{eq:objective_function}
\end{equation}
where, in the setting of linear mappings and neighborhoods with Gaussian falloffs,
$p_V$ is defined by (\ref{eq:neighborhood_probability_linear}) and is 
parameterized by the projection matrix $W_V$ of the linear mapping.

\textbf{Optimization.} The function $C(f_1,\ldots,f_{N_{\textrm{views}}})$ is 
a well-defined objective function for dependent projections and can be maximized
with respect to the mappings $f_V$ of each view; in the specific
case of linear subspaces, the objective function can be maximized with respect
to the projection matrices $W_V$ of the mappings. As the objective function is 
highly nonlinear with respect to the parameters, we use gradient based techniques
to optimize the projection matrices, specifically limited memory 
Broyden-Fletcher-Goldfarb-Shanno (L-BFGS).

\textbf{Initial penalty term.} Even with L-BFGS optimization, we have
empirically found that (\ref{eq:objective_function}) by itself can be hard to
optimize due to several local optima. To help reach a good local optimum, we use
iterative optimization over several rounds of L-BFGS with a shrinking additional
penalty: we add a data-driven penalty term to drive the objective away from the
worst local optima during the first rounds, and use the optimum found in each
round as initialization for the next. For the penalty we use the KL divergence
based dissimilarity between neighborhoods, summed over neighborhoods of all data
items $i$ and all pairs of views $(U,V)$, yielding
\begin{multline}
C_{\textrm{Penalty}}(f_1,\ldots,f_{N_{\textrm{views}}}) 
= \sum_{V=1}^{N_{\textrm{views}}} \sum_{U=1,U\neq V}^{N_{\textrm{views}}} \sum_{i=1}^{N_{\textrm{data}}}
(D_{KL}(p_{i,V},p_{i,U})+D_{KL}(p_{i,U},p_{i,V}))/2
\end{multline}
which is again a well defined function of the mapping parameters of each view
that can be optimized by L-BFGS. The KL divergence is useful as a penalty since
it heavily penalizes severe misses of neighbors (pairs $(i,j)$ where the
neighborhood probability is nonzero in one view but near-zero in another) and
hence drives the objective away from bad local optima; however, at the end of
optimization a small amount of misses must be tolerated since views may not
fully agree even with the best mapping, hence at the end of optimization the
original objective (\ref{eq:objective_function}) is a more suitable choice. We
will thus shrink away the KL divergence penalty during optimization.
The objective function with the added penalty then becomes
\begin{equation}
C_{\textrm{Total}}(f_1,\ldots,f_{N_{\textrm{views}}})
C(f_1,\ldots,f_{N_{\textrm{views}}}) - \gamma C_{\textrm{Penalty}}(f_1,\ldots,f_{N_{\textrm{views}}}) 
\end{equation}
where $\gamma$ controls the amount of penalty. We initially set $\gamma$ so that
the two parts of the objective function are equal for the initial
mappings, $C(f_1,\ldots,f_{N_{\textrm{views}}}) = \gamma
C_{\textrm{Penalty}}(f_1,\ldots,f_{N_{\textrm{views}}})$, and shrink $\gamma$
exponentially towards zero with respect to the L-BFGS rounds; in experiments
we multiplied $\gamma$ by a multiplier $0.9$ at the start of each 
L-BFGS round to accomplish the exponential shrinkage.

\textbf{Time complexity}. The proposed method needs to calculate the neighbor distributions for all views, 
and optimizes the objective function involving each pairs of views, and thus
the naive implementation requires $O(dN_{data}^2N_{views}^2)$ of time, where $d$ is the maximum number of dimensions
among all views.
We are aware of several acceleration techniques \citep{yang2013scalable, vandermaaten2014, vladymyrov2014linear}
for doing neighbor embedding, which can reduce the time complexity of a single view
from $O(N_{data}^2)$ to $O(N_{data}\log N_{data})$ or even $O(N_{data})$. 
But scalability is not our first concern in this paper, so we use the naive $O(N_{data}^2)$ implementation 
for calculating the neighbor distributions for each view involved.



\section{Properties of the Method and Extensions}

\textbf{Information retrieval.} Our cost function measures success in a neighbor
retrieval task of the analyst: we maximize the count of retrieved 
true neighbors across views, and initially penalize by the severity of misses
across views.\\
\textbf{Invariances.} For any subspace of any view (input space),
the neighborhood probabilities
(\ref{eq:neighborhood_probability_general}) and (\ref{eq:neighborhood_probability_linear})
depend only on pairwise distances and are thus invariant to global translation, rotation, and mirroring of input data in that
subspace. Therefore the cost function is also invariant to differences of
global translation, rotation, and mirroring between views and is able to discover
dependencies of views despite such differences. The invariance is even stronger:
for a given subspace in each view, if any subset of data items is isolated from 
the rest in all views, such that the
neighborhood probability is zero for any data pair $(i,j)$ where one items 
is in the subset and the other is outside, then the same 
invariances to translation, rotation and mirroring applies to each such isolated
subset separately, as long as the the translations, rotations, and mirrorings 
preserve the isolation of the subsets.\\ 
\textbf{Dependency is measured between subspaces as a whole.}
Unlike CCA where each canonical component of one view has a particular correlated
pair in the other view, in our method dependency is maximized with respect to the
entire subspaces (transformed representations) of each view as a whole, since
neighborhoods of data depend on all coordinates within the dependent subspace.
Our method hence takes into account within-view feature dependencies
when measuring dependency. Furthermore, dependent subspaces do not even need
to be same-dimensional, and in some views we can choose not to reduce dimensionality
at all but to learn a metric (full-rank linear transformation).\\
\textbf{Finding dependent neighborhoods between feature-based views 
and views external neighborhoods.}
We point out that in some domains, a subset of the available data views may
directly provide neighborhood relationships or similarities between data items,
such as known friendships between people in a social network, known
followerships between Twitter users, or known citations between scientific
papers. If available, such relationships or similarities can directly be used in
place of the feature-based neighborhood probabilities $p_V(j|i;f_V)$ discussed 
above. This presents an interesting similarity to previous method \citep{Peltonen09wsom} 
which was used to find similarities of one view to an external neighborhood definition;
our present method contains this task as one special case.\\
\textbf{Alternative mathematical forms.}
It is simple to replace the exponential falloff in 
(\ref{eq:neighborhood_probability_general}) and (\ref{eq:neighborhood_probability_linear})
with another type of falloff if appropriate for the data domain. For
example, some dimensionality reduction methods have used a t-distributed
definition for data neighborhoods \citep{Venna10jmlr,vanderMaaten08}. Such a 
replacement preserves the invariances discussed above.\\
\textbf{Alternative forms of the transformations.}
In place of linear transformations one can substitute
another parametric form such as a neural network, and optimize the objective
function with respect to its parameters; the transformation can be
chosen on a view-by-view basis. Difficulty of the
nonlinear optimization may differ for different transformations and the best
form for more general nonlinear transformations is outside the scope of the
current paper.


\section{Experiments}
We demonstrate the neighborhood preservation ability of our method on an artificial data set with multiple dependent groups between views, 
and three real data sets, including a variant of MNIST handwritten digit database \citep{mnistlecun}, 
Wisconsin X-ray Microbeam Database \citep{xrmb}, and a cell cycle-regulated genes data set \citep{spellman-cellcycle}. 
We compare our method with CCA. On the artificial data set, we measure the performance by a correspondence between the 
found projection components and the known ground truth. On the real data sets, we measure the performance by the mean precision-mean recall curve
on the side with a smaller number of retrieved neighbors.

\subsection{Experiment on artificial data sets}
We generate an artificial data set with 2 views with 1000 data points and 5 dimensions each. The dimensions are built iteratively by
creating a pair of dimensions with multiple dependent groups, and assigning one dimension to view 1, and the other to view 2,
detailed as follows. 

Let $X^{(1)}, X^{(2)}\in \mathbb{R}^{5\times 1000}$ be view 1 and view 2 respectively, and $X^{(V)}_i$ be the $i$-th dimension 
in view $V$ ($i\in\{1,\cdots,5\},V\in\{1,2\}$). We call $(X^{(1)}_i, X^{(2)}_i)$ the $i$-th dimension pair. For each $i$, 
we create 20 groups $\{g_{i,1},\cdots,g_{i,20}\}$, each of which has 50 data points. For each $g_{ij}\ (j\in\{1,\cdots,20\})$, 
we first sample a group mean $m_{ij}^{(V)}\sim \mathcal{N}(0, 5)$ for each view $V$, and give it a common perturbation 
$\epsilon_{ijk}\sim U[-0.5, 0.5]\ (k\in\{1,\cdots,50\})$ for each point in the group across the views. 
Let $\hat{x}_{ijk}^{(V)}\triangleq Fm_{ij}+\epsilon_{ijk}$, where $F\in \{-1, 1\}$ is a random variable for flipping,
allowing the data set to have positive or negative correlation inside the group. 
After collecting the generated $\hat{x}_{ijk}^{(1)}$ and $\hat{x}_{ijk}^{(2)}$ into matrices $\hat{X}^{(1)}, \hat{X}^{(2)}\in \mathbb{R}^{5\times 1000}$,
we randomly permute data points in the same way for $\hat{X}^{(1)}_i$ and $\hat{X}^{(2)}_i$, but differently for different $i$,
to ensure the independency between dimensions. Finally we perform a PCA between $\hat{X}^{(1)}_i$ and $\hat{X}^{(2)}_i$ for each $i$, 
to remove the correlation inside the dimension pair, and get the final $X^{(1)}_i$ and $X^{(2)}_i$, and subsequently $X^{(1)}$ and $X^{(2)}$. 

\begin{figure}[!htb]
\centering
\vspace{-3mm}
\includegraphics[width=0.32\textwidth]{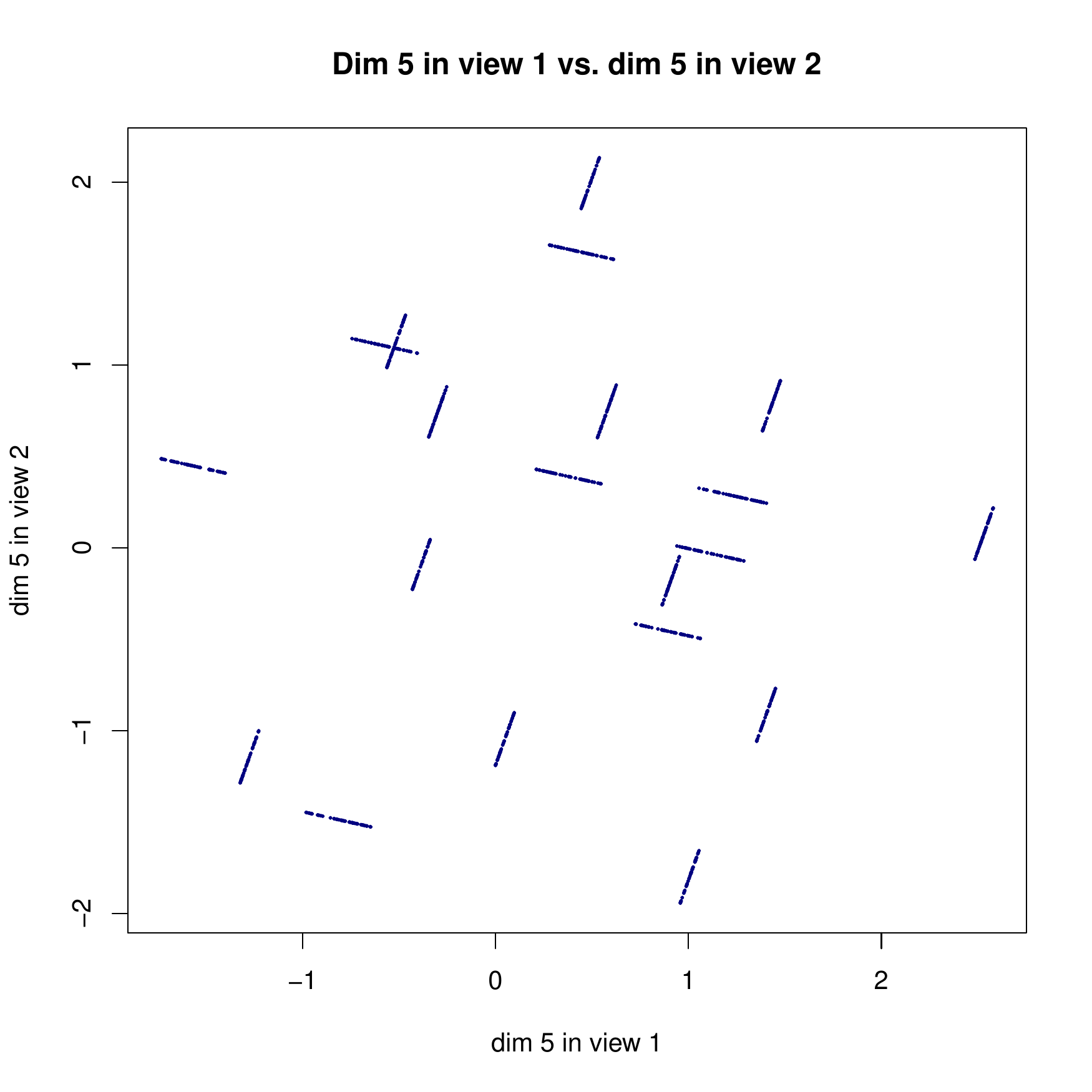}
\includegraphics[width=0.32\textwidth]{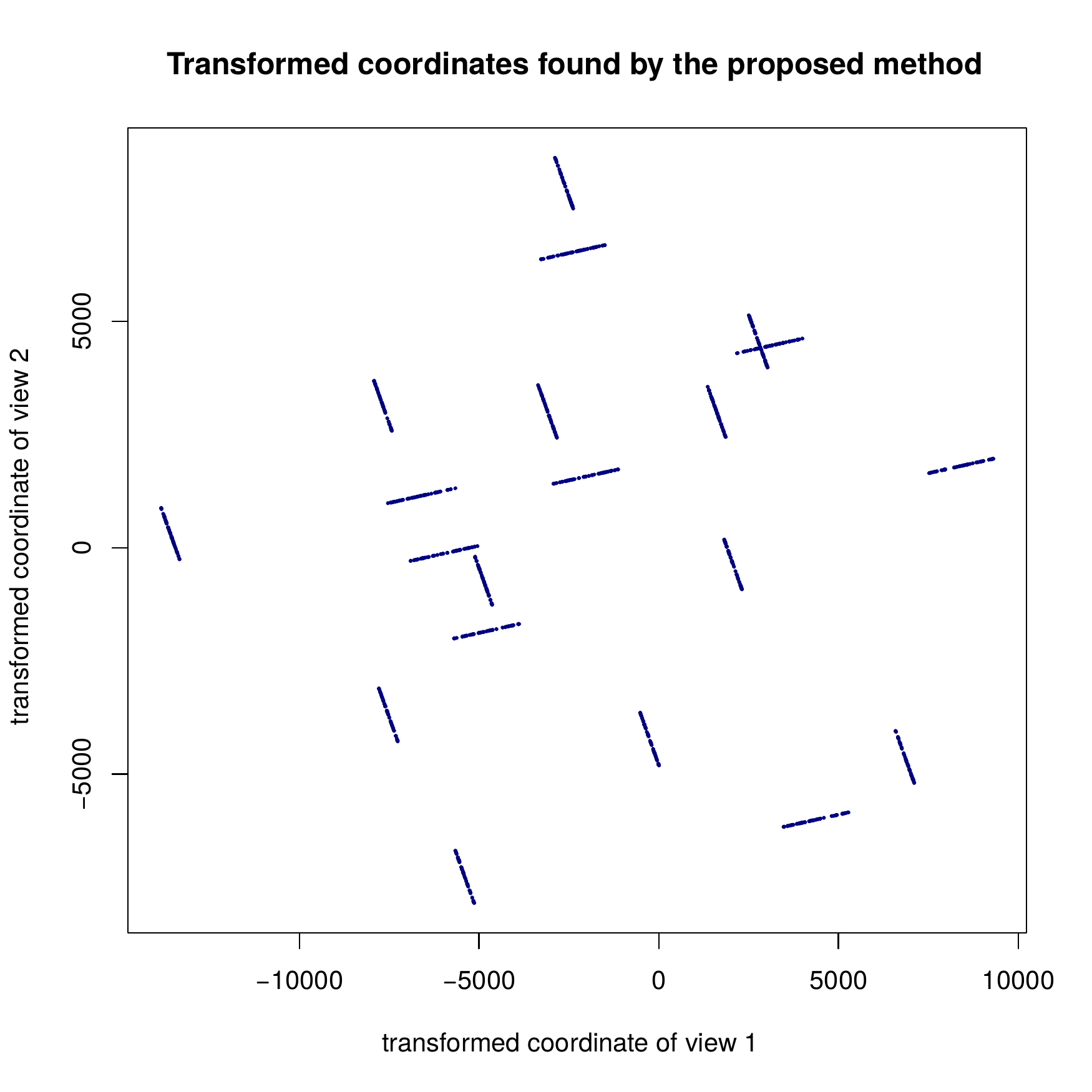}
\includegraphics[width=0.32\textwidth]{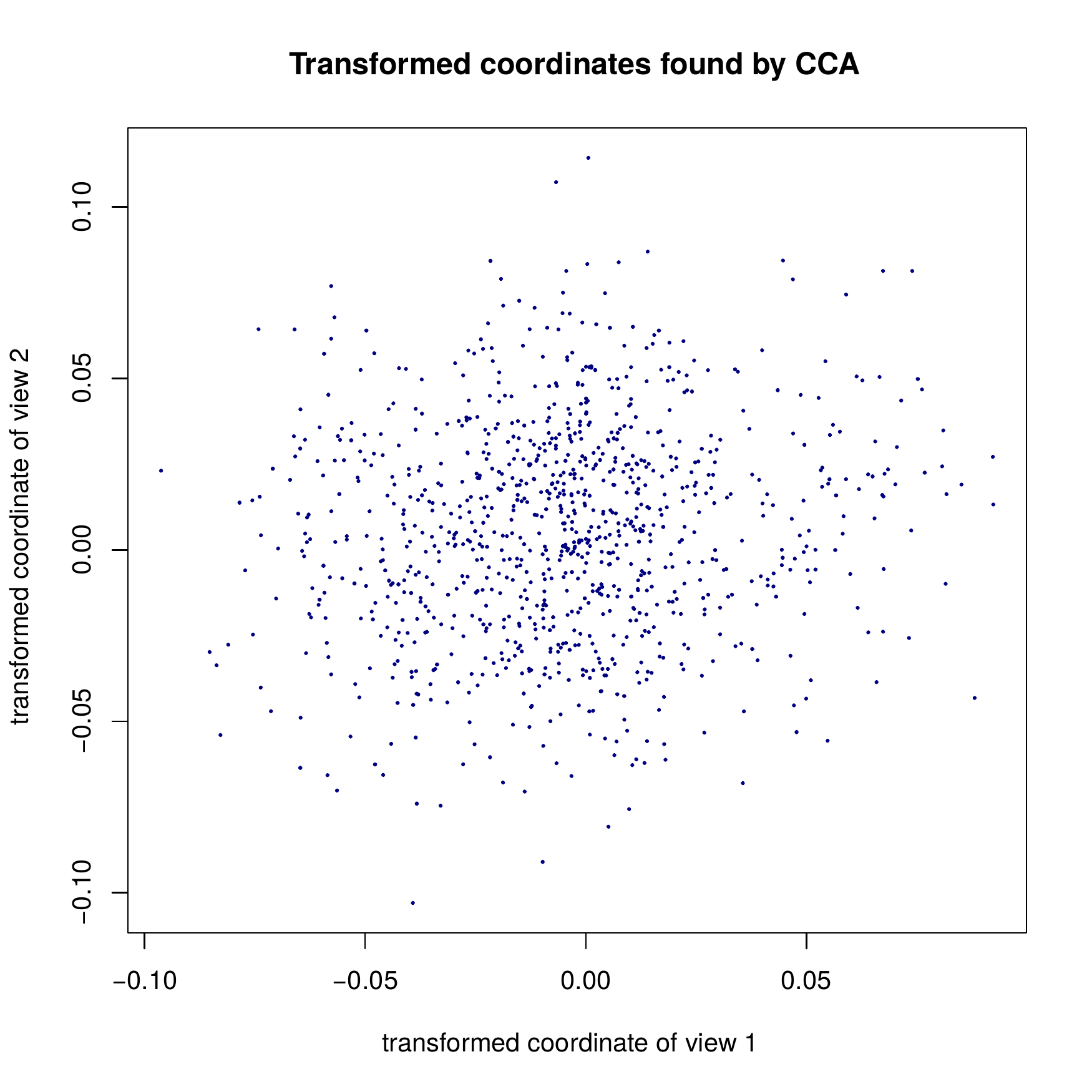}
\vspace{-3mm}
\caption{The generated artificial data has 5 dimensions in each view and 20 dependent groups in each dimension. 
\textbf{Left}: the 5th dimension in view 1 vs. the 5th dimension in view 2, 
as an example of the dependency in a certain dimension pair between the two views.
\textbf{Middle}: transformed coordinates of view 1 vs. transformed coordinates of view 2 from our method.
It recovers the dependency between the two views in the 5th dimension despite of the mirroring and the scale.
\textbf{Right}: transformed coordinates of view 1 vs. transformed coordinates of view 2 from CCA, where the dependency cannot be seen.}
\label{fig:toy-data}
\end{figure}

We pursue 2 transformations mapping from the 5-dimensional original space to a 1-dimensional latent space for the two views.
In this case, the ground truth transformations for both views will be a linear projection $W^{(i)}=(0,\cdots,0,1,0,\cdots,0)\in\mathbb{R}^{1\times 5}$ 
where the 1 appears only at the $i$-th position.
Results are shown in Fig. \ref{fig:toy-data}: compared with CCA, our method successfully finds one of the ground truth transformations (= the 5th one),
despite of the mirroring and the scale, recovering the dependency between the two views.

We measure the performance by the correspondence between the found projections and the ground truth transformation as defined below. 
Let $W_1, W_2\in\mathbb{R}^{1\times 5}$ be the found projections from either method, define
\begin{equation}
\label{eq:corr-measure-toy-data}
\mathrm{Corr}(W_1, W_2)=\max_i \frac{1}{2}\left(\frac{|W^{(i)}W_1^\mathrm{T}|}{\|W_1\|_2}+\frac{|W^{(i)}W_2^\mathrm{T}|}{\|W_2\|_2}\right)
\end{equation}
as the correspondence score. A high score indicates a good alignment between the found projections and the ground truth.
We repeat the experiment and calculate the correspondence on 20 artificial data sets generated in the same way, 
and summarize the statistics in Table \ref{tab:toy-data}. Our method outperforms over CCA by successfully finding the dependency within all 20 artificial data sets.

\begin{table}[!htb]
\begin{center}
\vspace{-3mm}
\begin{tabular}{|c|c|c|}
\hline
& Mean & Std \\
\hline
\hline
Our method & 1.00 & 0.00 \\
\hline
CCA & 0.51 & 0.043 \\
\hline
\end{tabular}
\vspace{-3mm}
\end{center}
\caption{The means and the standard deviations of the correspondence measure as defined in Eq. \eqref{eq:corr-measure-toy-data}
from our method and CCA. Our method clearly outperforms by successfully recovering the dependency in all artificial data sets.}
\label{tab:toy-data}
\end{table}

\subsection{Experiment on real data sets}

In this experiment we show our method helps match neighbors between the subspaces of two views after transformation.
We use the following 3 data sets for the demonstration.

\textbf{MNIST handwritten digit database} (MNIST). MNIST contains gray-scale pixel values from images
with size $28\times 28$ of 70000 hand-written digits (60000 in the training set, 10000 in the testing set).
We randomly choose 100 images for each digit from the training set, take the left half of each image as view 1, and the right half as view 2,
by which we have two 392-dimensional data matrices as the views, consisting of 1000 samples each. 
The testing set also has 1000 random samples.

\textbf{Wisconsin X-ray Microbeam Database} (XRMB). XRMB contains two views with simultaneous speech and tongue/lip/jaw 
movement information from different speakers. Specifically, we use the preprocessed version as in \cite{LopezPaz2014} 
\footnote{Data files available at \url{https://github.com/lopezpaz/randomized_nonlinear_component_analysis}},
where the acoustic features are the concatenation of 
mel-frequency cepstral coefficients (MFCCs) \citep{davis80} of consecutive frames from the speech, 
with 273 dimensions at each time point, 
and the articulatory features are the concatenation of continuous tongue/lip/jaw displacement measurements from the same frames,
with 112 dimensions at each time point. We randomly choose 2000 samples out of 50000 from the original data for training, 
and another 2000 random samples for testing.

\textbf{Cell Cycle–regulated Genes} (Cell-cycle). The cell-cycle data are from two different experiment measurements
of cell cycle–regulated gene expressions for the same set of 5670 genes from \cite{spellman-cellcycle}. 
We choose the measurements from experiments ``$\alpha$ factor arrest'' and ``Cln3'', 
preprocess the data as in \cite{tripathi08}. We randomly take a subset of 4536 samples for training, and the rest 1134 samples for testing.

\begin{figure}[!htb]
\centering
\begin{tabu}{c@{\hskip3pt}c@{\hskip3pt}c@{\hskip3pt}c@{\hskip3pt}c}
& \multicolumn{2}{c}{Training} & \multicolumn{2}{c}{Testing} \\
\hline\vspace{-3mm}\\
\rowfont{\scriptsize} & \parbox{2.5cm}{\centering{view 1 vs. transformed coordinates of view 2}} &
\parbox{2.5cm}{\centering{transformed coordinates of view 1 vs. transformed coordinates of view 2}} &
\parbox{2.5cm}{\centering{view 1 vs. transformed coordinates of view 2}} &
\parbox{2.5cm}{\centering{transformed coordinates of view 1 vs. transformed coordinates of view 2}} \vspace{2mm}\\
\hline\\
\raisebox{1.5\normalbaselineskip}[0pt][0pt]{\rotatebox[origin=l]{90}{MNIST}} &
\includegraphics[width=0.24\textwidth]{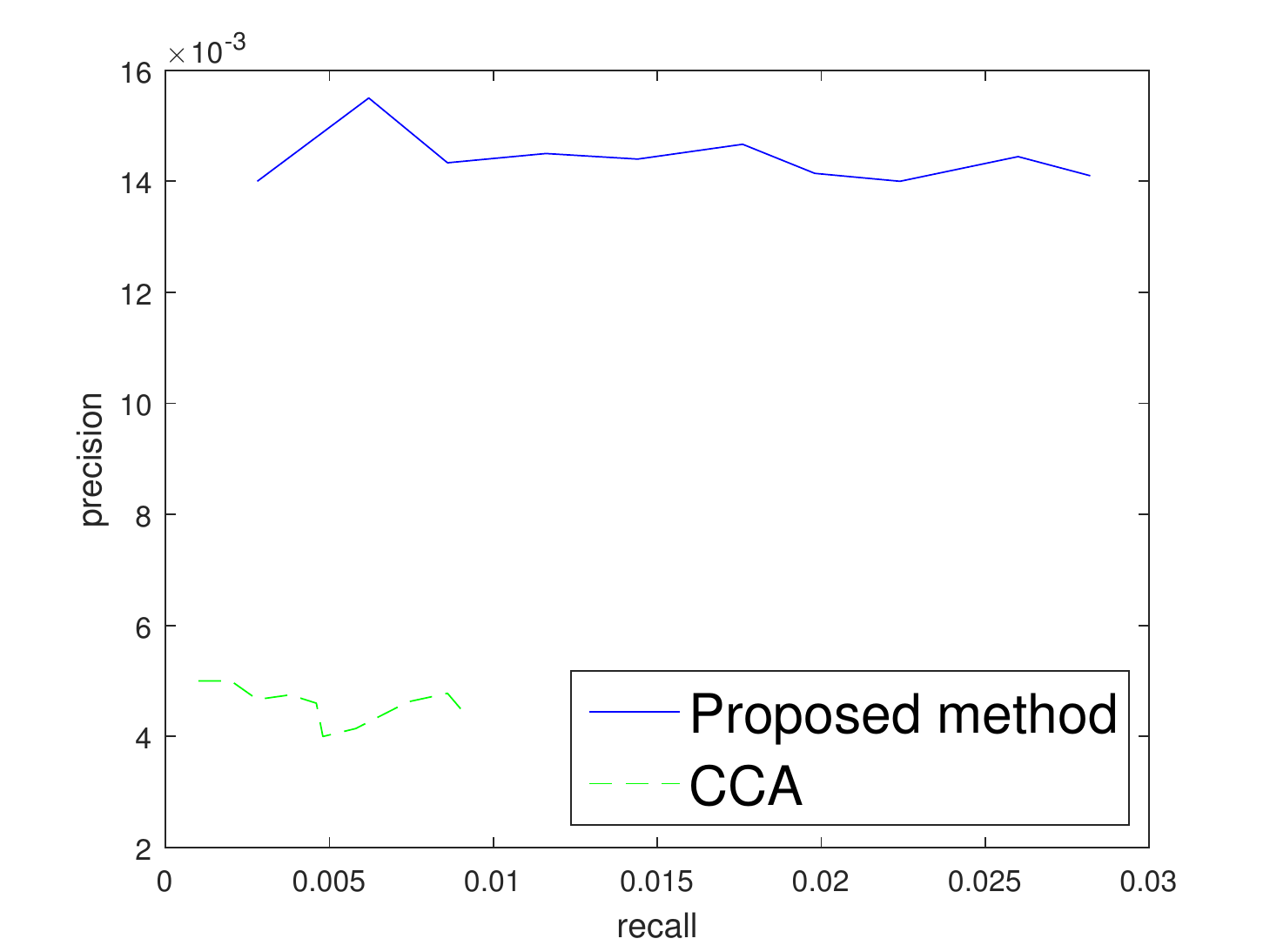} &
\includegraphics[width=0.24\textwidth]{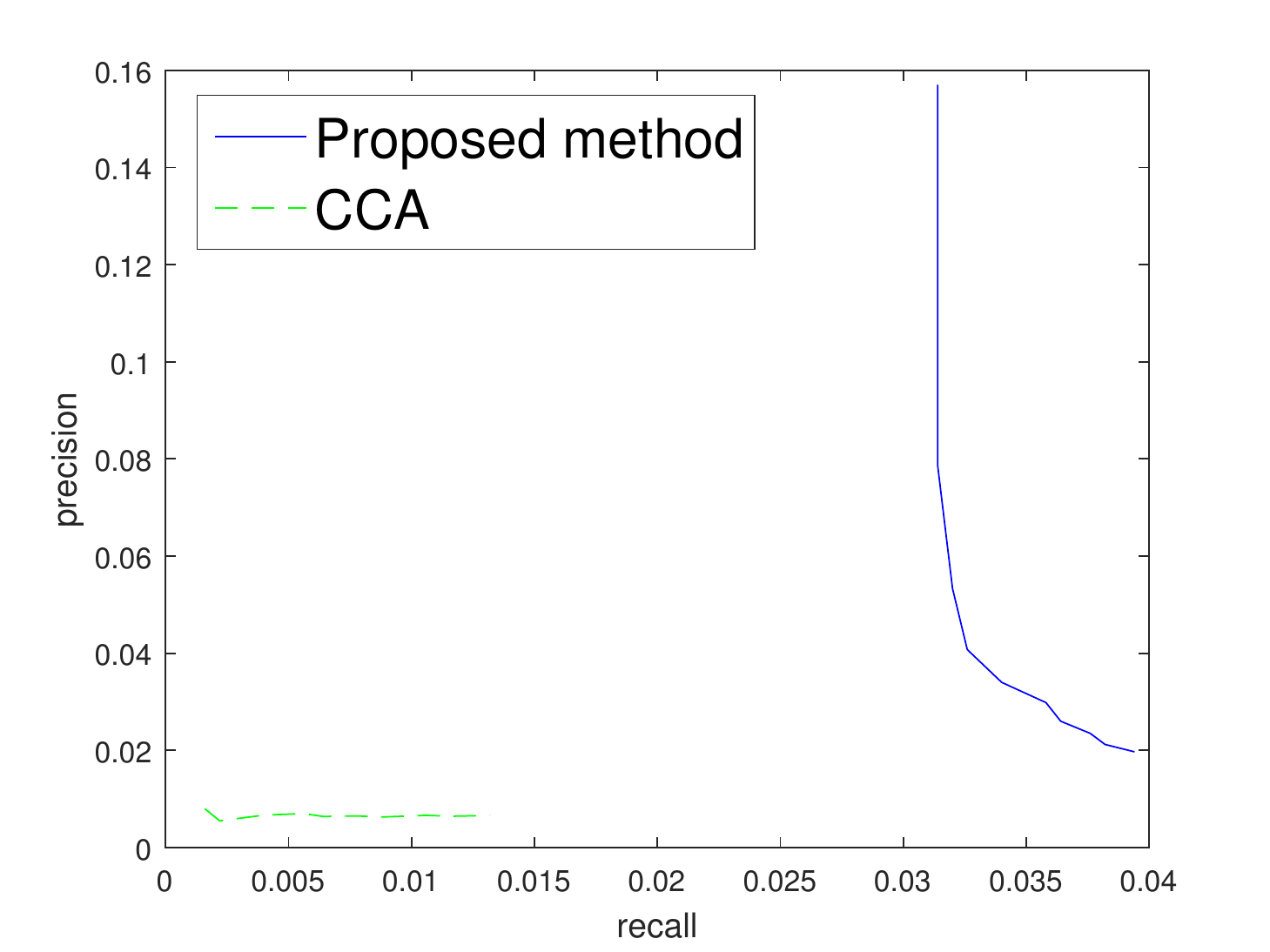} &
\includegraphics[width=0.24\textwidth]{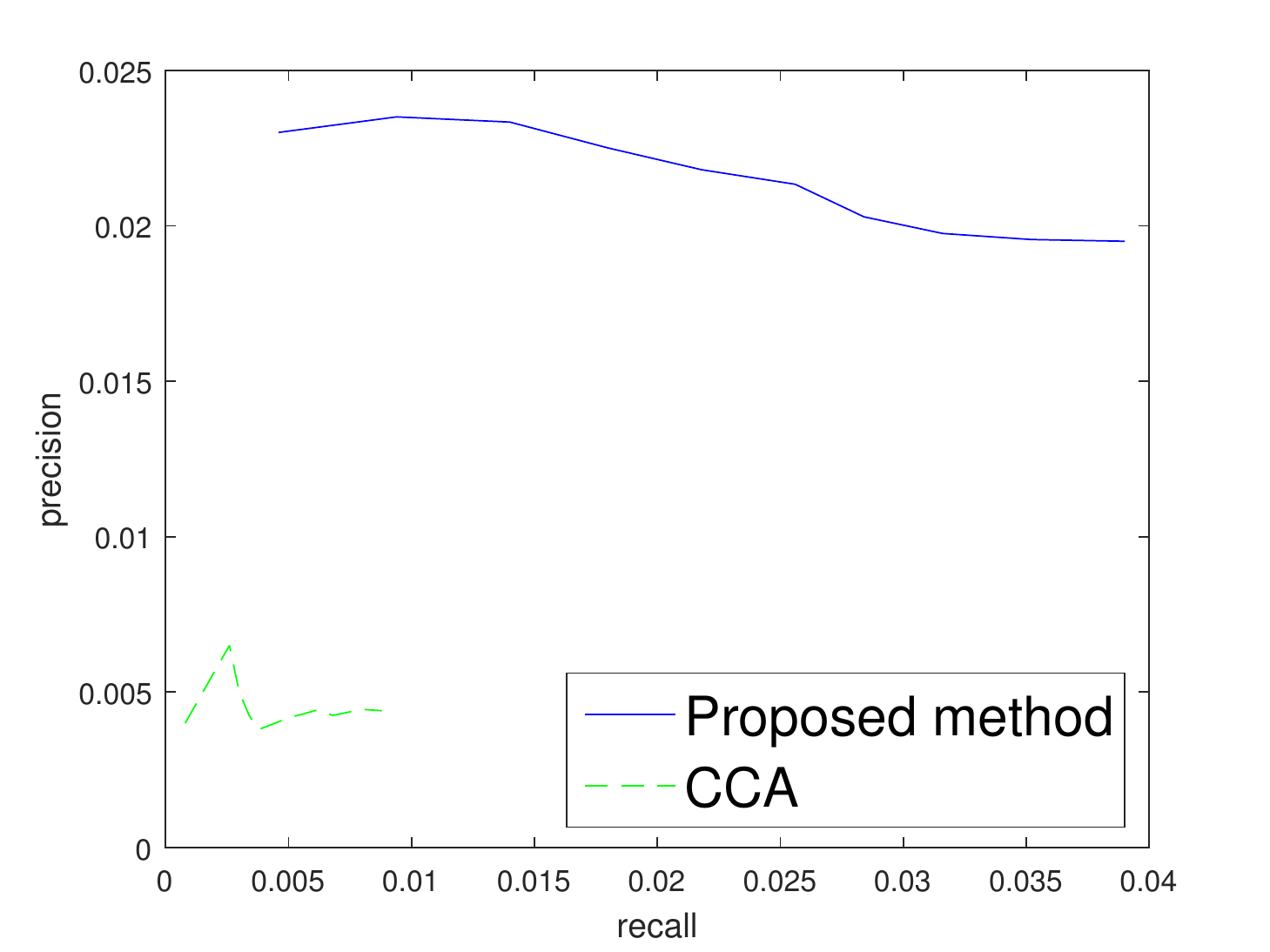} &
\includegraphics[width=0.24\textwidth]{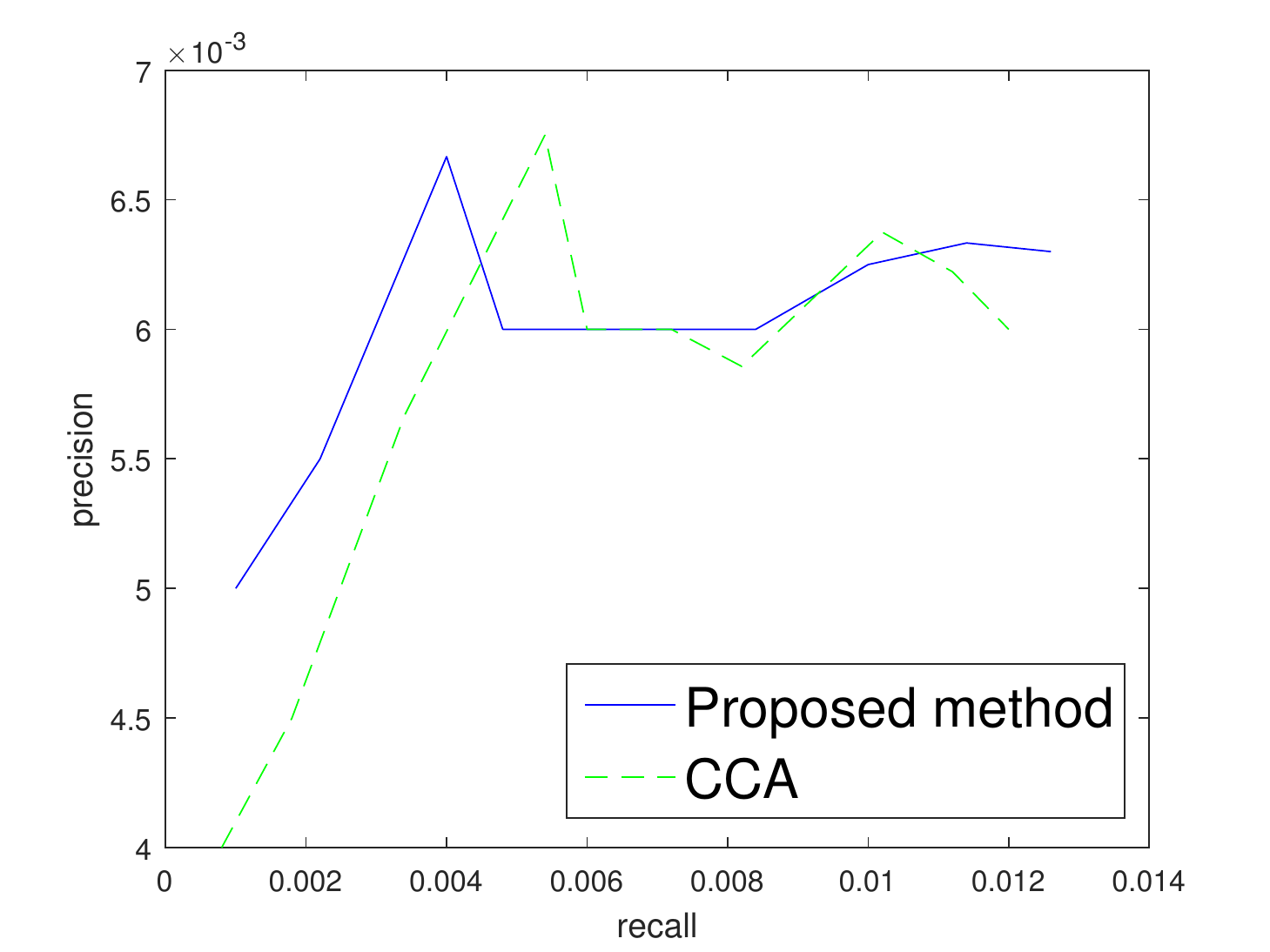} \vspace{-3mm}\\
\raisebox{1.5\normalbaselineskip}[0pt][0pt]{\rotatebox[origin=l]{90}{XRMB}} &
\includegraphics[width=0.24\textwidth]{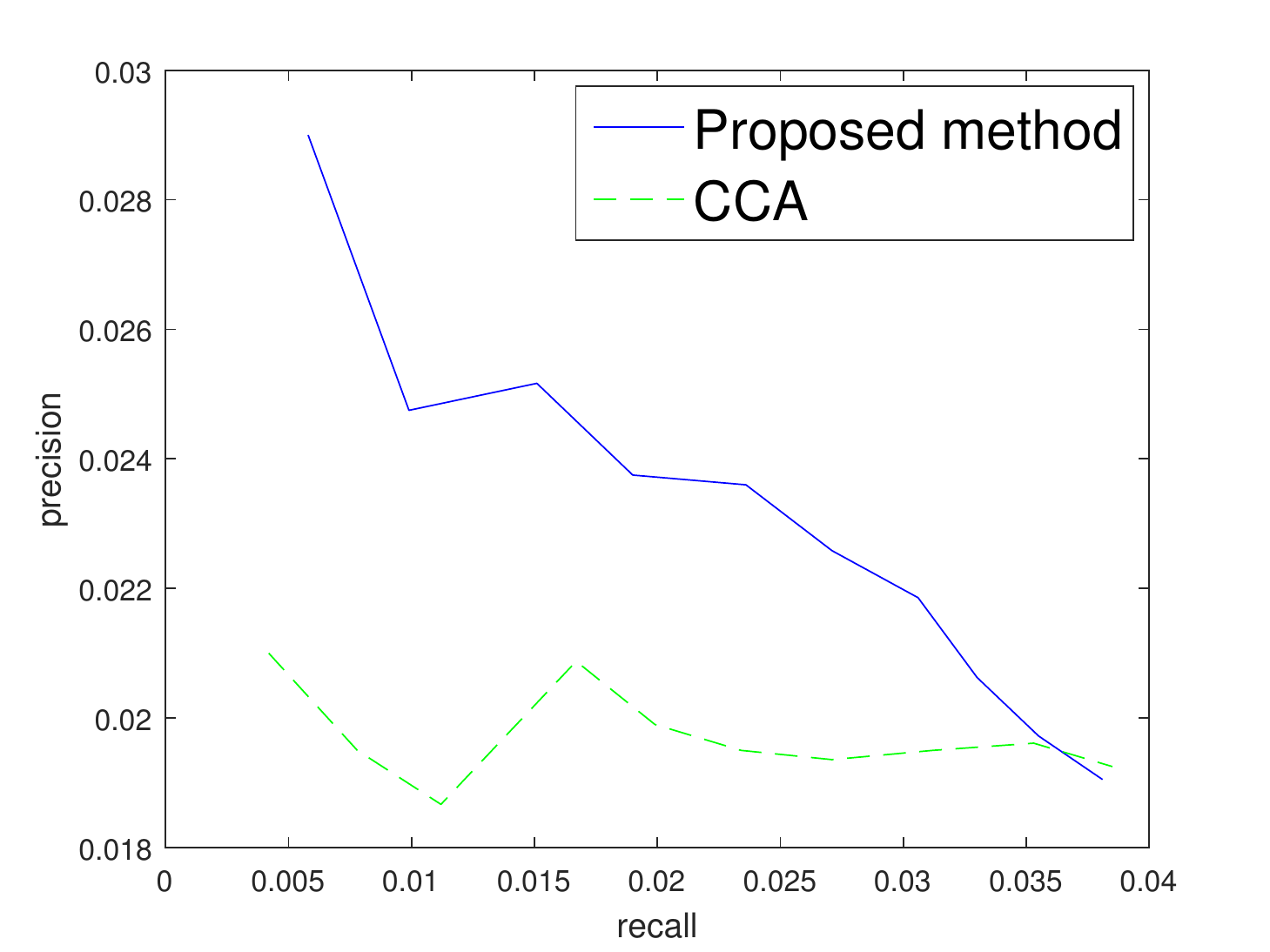} &
\includegraphics[width=0.24\textwidth]{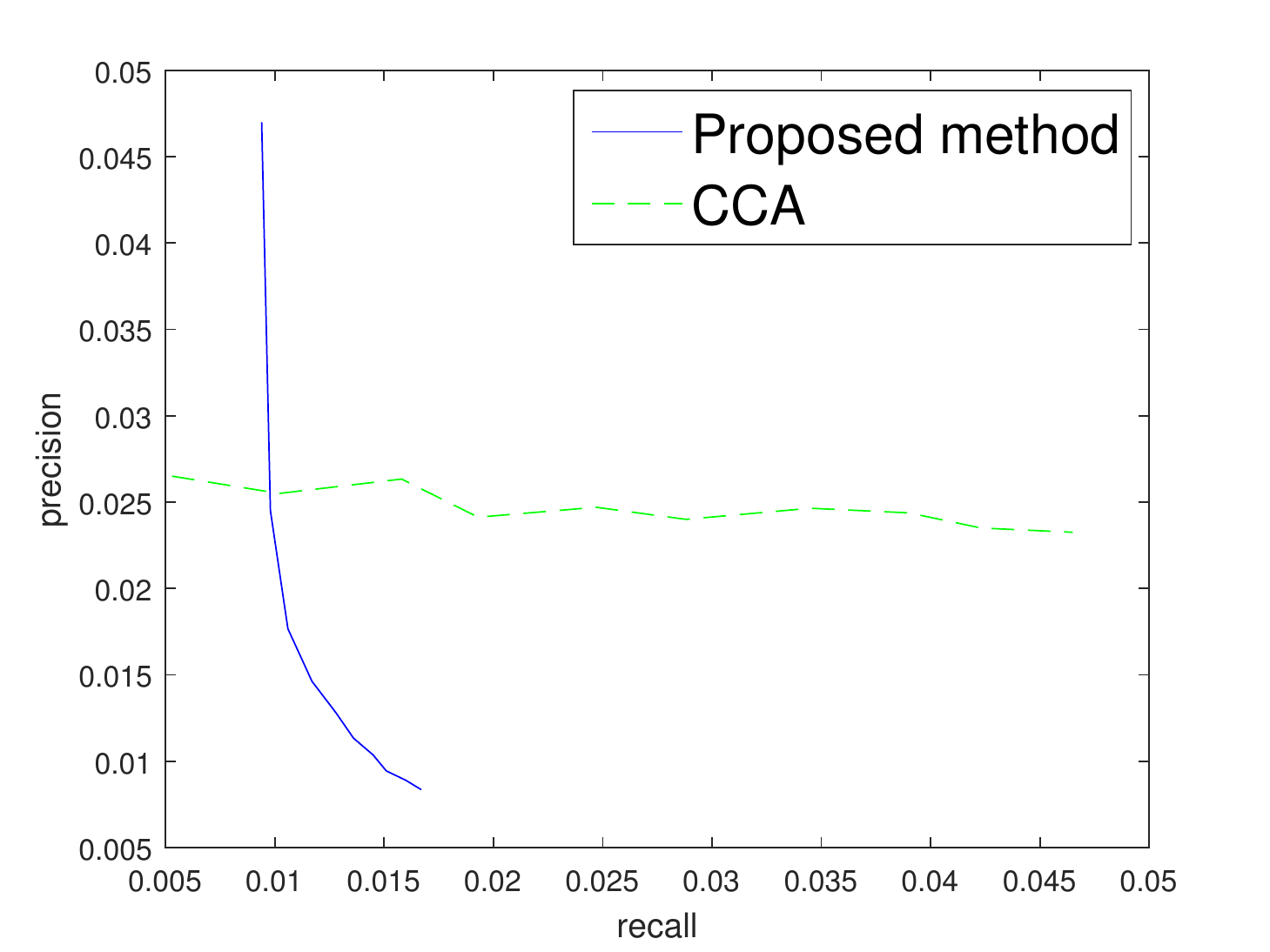} &
\includegraphics[width=0.24\textwidth]{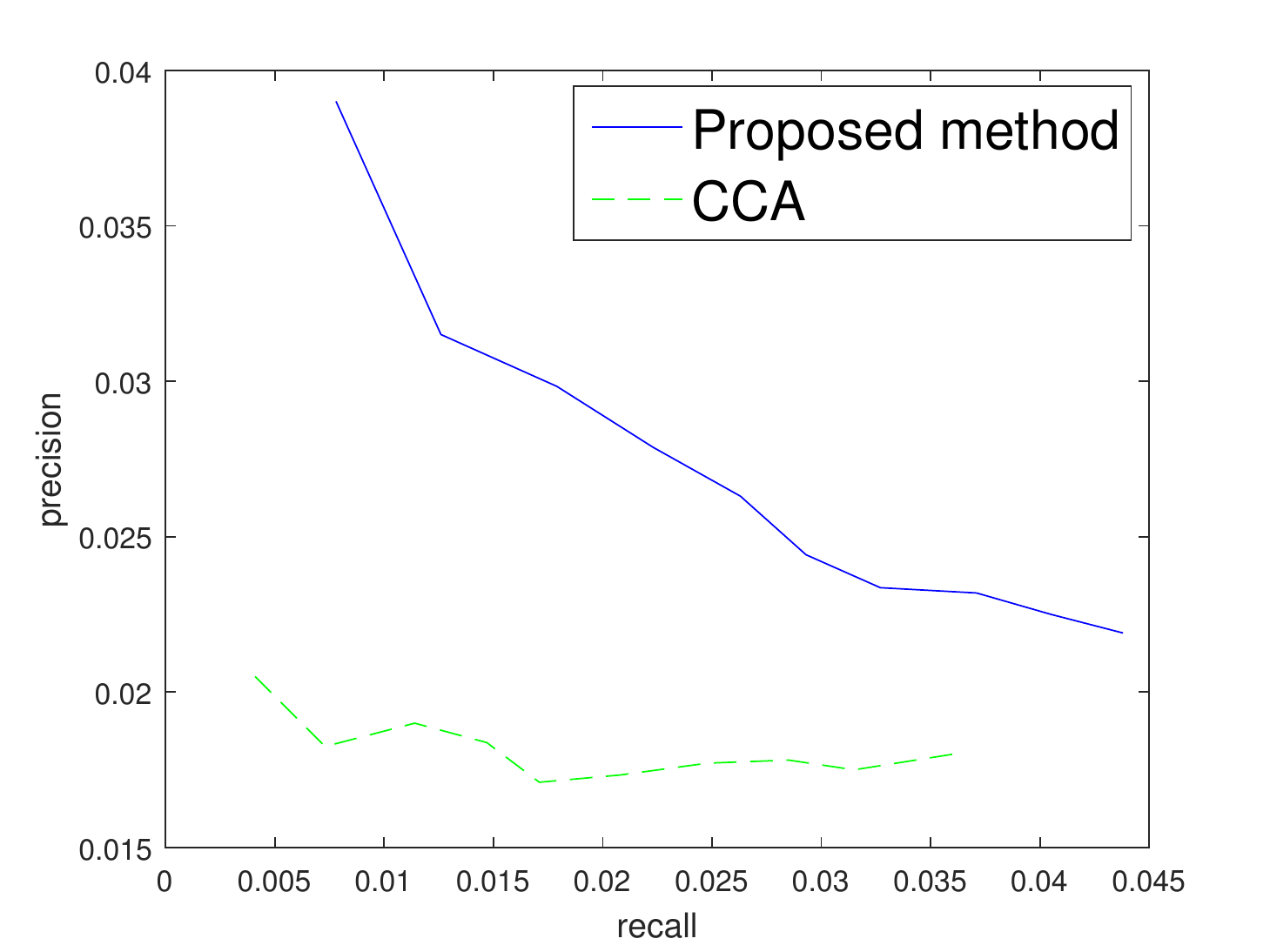} &
\includegraphics[width=0.24\textwidth]{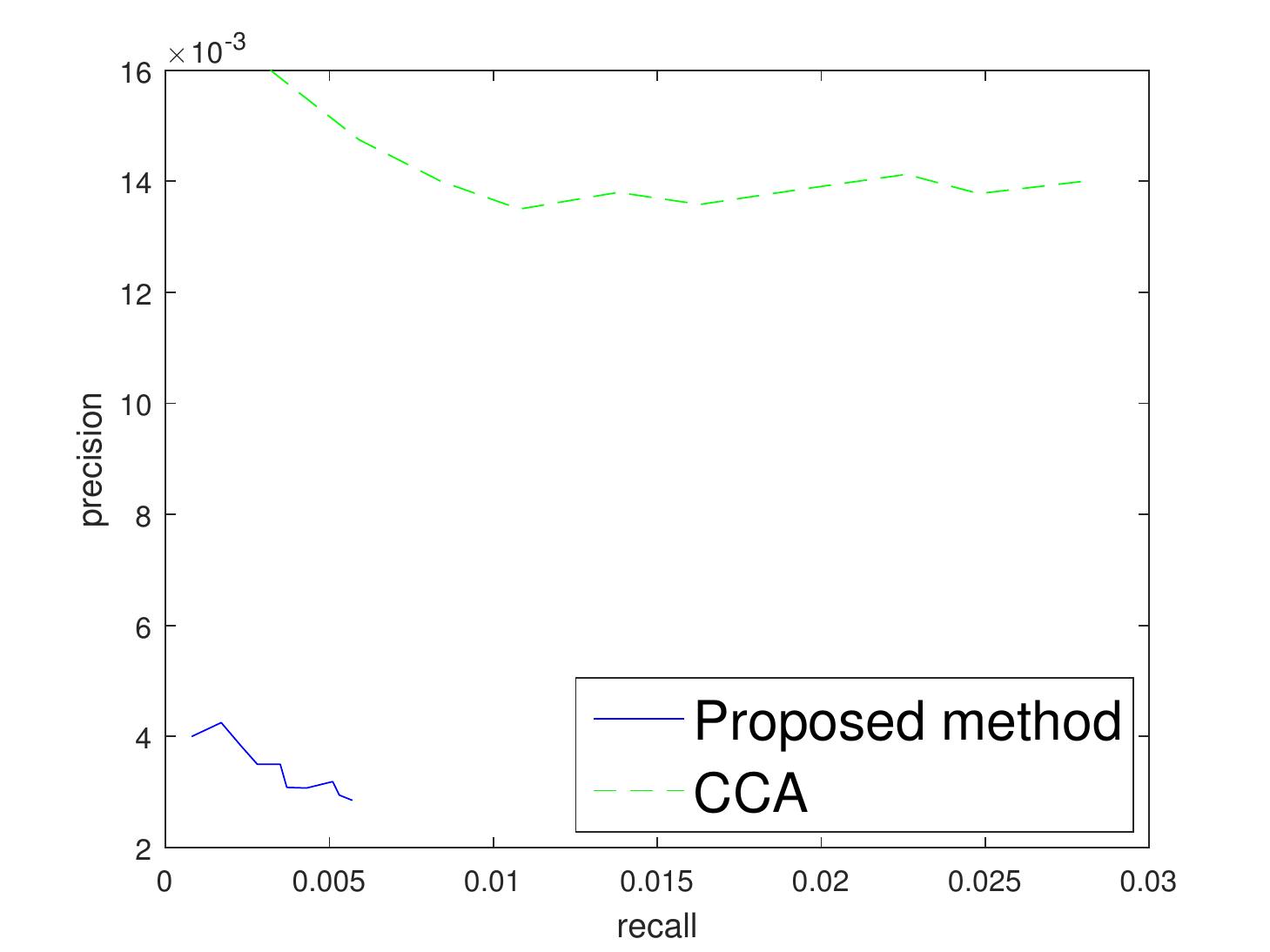} \vspace{-3mm}\\
\raisebox{\normalbaselineskip}[0pt][0pt]{\rotatebox[origin=l]{90}{Cell-Cycle}} &
\includegraphics[width=0.24\textwidth]{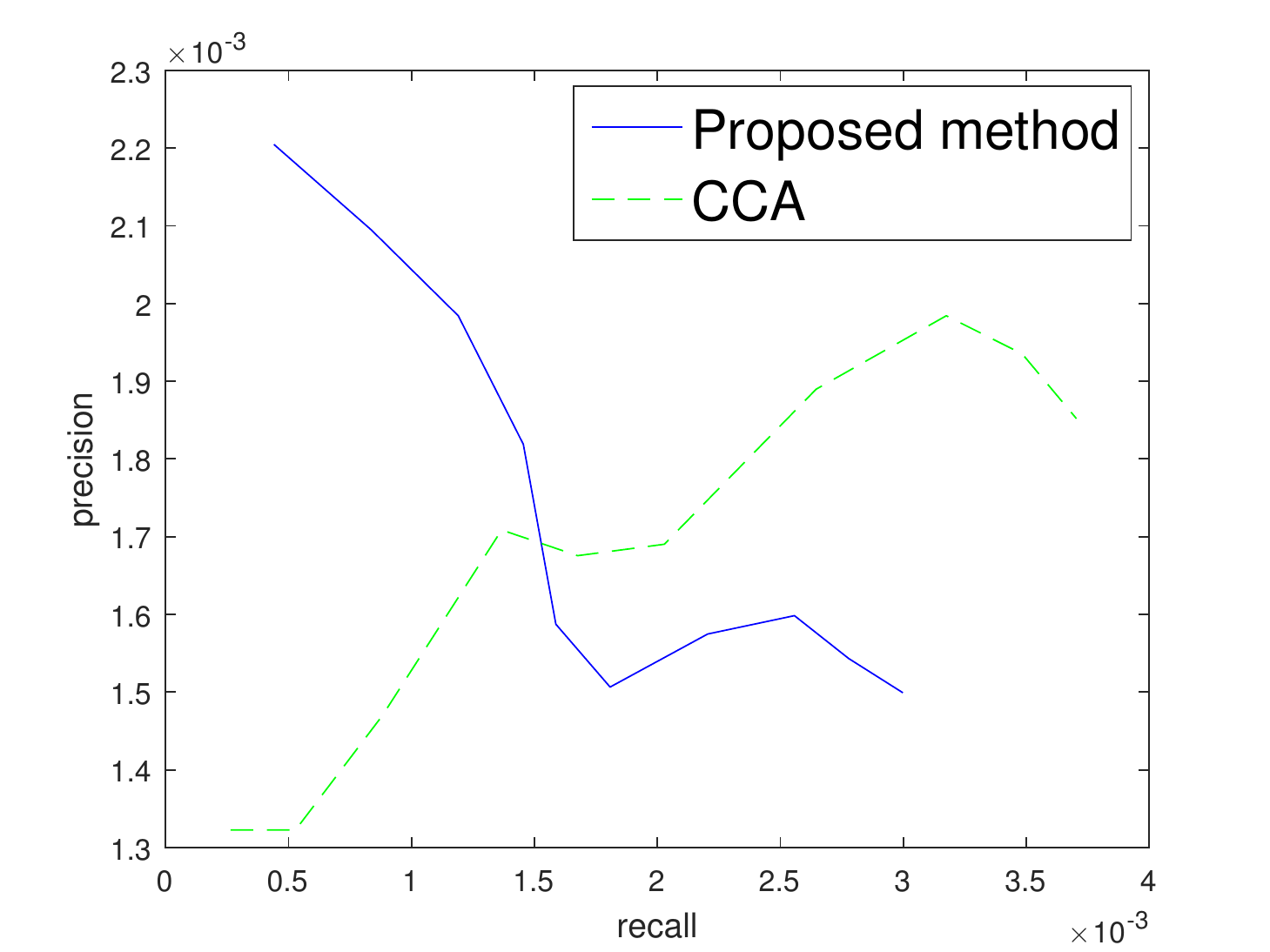} &
\includegraphics[width=0.24\textwidth]{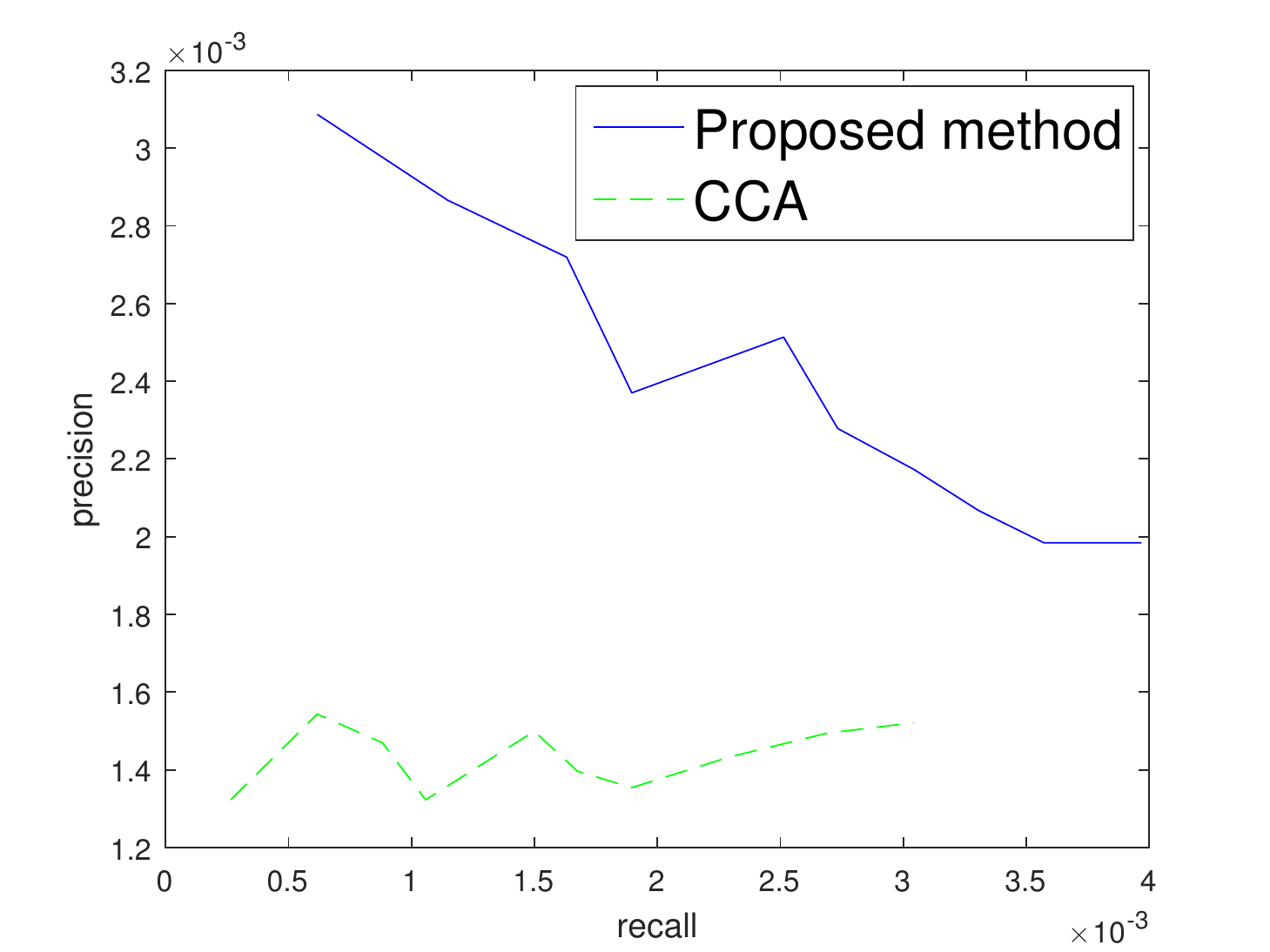} &
\includegraphics[width=0.24\textwidth]{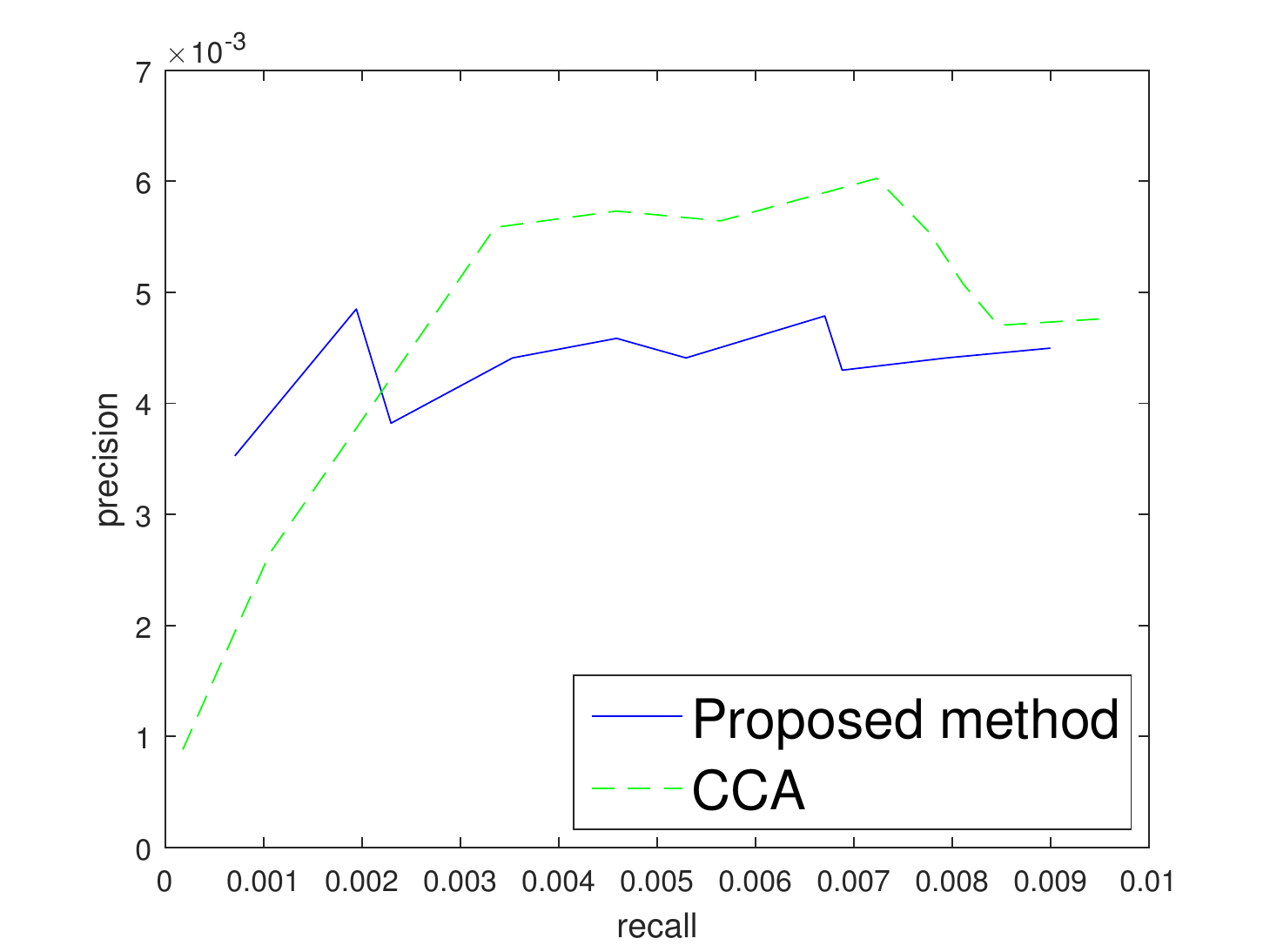} &
\includegraphics[width=0.24\textwidth]{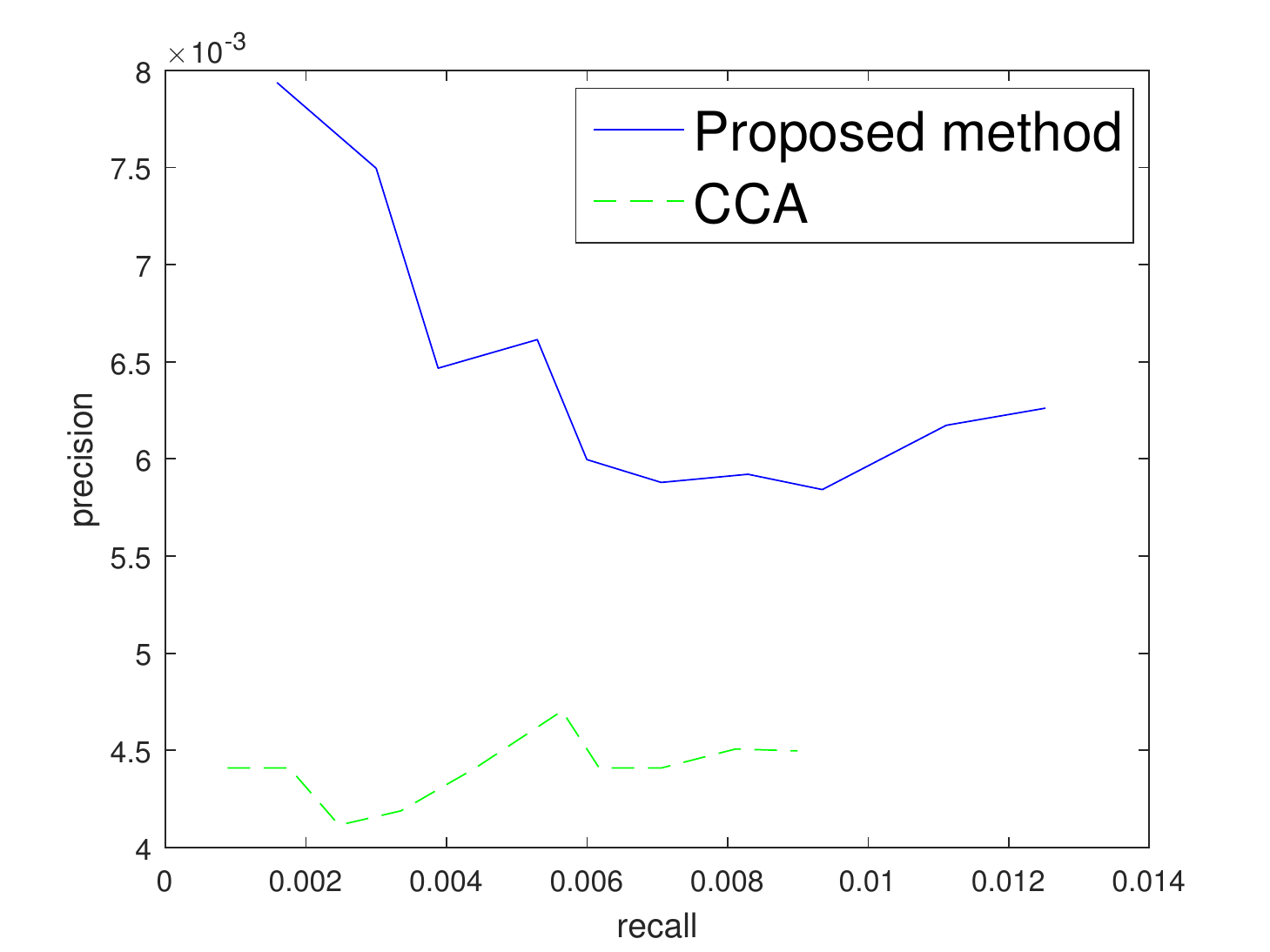}
\end{tabu}
\caption{The mean precision-mean recalls curves from different real data sets.
\textbf{Left 6}: the curves from the training sets. 
\textbf{Right 6}: the curves from the testing sets. 
\textbf{Column 1, 3}: the curves with view 1 as the ground truth. 
\textbf{Column 2, 4}: the curves with the subspace from view 1 as the ground truth.
Our method performs better than CCA as the curves from our method are mostly
located at the top and/or right to the curves from CCA, especially when the number of retrieved neighbors
from the subspace of view 2 is small. The figure from the XRMB testing set 
with the subspace from view 1 as the ground truth is the only exception, meaning that over-learning may have occurred.}
\label{fig:real-data}
\end{figure}

For the above data sets, we pursue a pair of transformations mapping onto 2-dimensional subspaces for the views. We measure the performance by the 
mean precision-mean recall curves between 1) the two subspaces from the transformations, and 
2) one of the original views and the subspace from the transformation for the other view. 
The curve from the method with a better neighbor matching will be at the top and/or right side in the figure, 
indicating that the method has achieved better mean precision and/or better mean recall.
We set the number of neighbors in the ground truth as 5, and let the number of 
retrieved neighbors varies from 1 to 10 since we focus on the performance of the matching for the nearest neighbors.
Fig.\ref{fig:real-data} shows the resulting curves. Our method outperforms CCA since the curves from our method mostly
are located at top and/or right to the curves from CCA. 
It worths pointing out that our method outperforms in preserving the neighborhood
relation not only across the two dependent subspaces that we optimize for, but also between the one of the original view
and the subspaces of the other view, especially when the number of retrieved neighbors is small (the left side of the figures).

\section{Conclusion and discussion}
We have presented a novel method for seeking dependent subspaces across
multiple views based on preserving neighborhood relationships between data
items. The method has strong invariance properties, detects nonlinear dependencies,
is related to an information retrieval (neighbor retrieval) task of the analyst,
and performs well in experiments.

\bibliography{cca_nr}
\bibliographystyle{iclr2016_conference}

\end{document}